# A Model-Predictive Motion Planner for the IARA Autonomous Car*

Vinicius Cardoso, Josias Oliveira, Thomas Teixeira, Claudine Badue, Filipe Mutz,
Thiago Oliveira-Santos, Lucas Veronese, Alberto F. De Souza, *Senior Member, IEEE*

*Abstract*—We present the Model-Predictive Motion Planner (MPMP) of the Intelligent Autonomous Robotic Automobile (IARA). IARA is a fully autonomous car that uses a path planner to compute a path from its current position to the desired destination. Using this path, the current position, a goal in the path and a map, IARA's MPMP is able to compute smooth trajectories from its current position to the goal in less than 50 ms. MPMP computes the poses of these trajectories so that they follow the path closely and, at the same time, are at a safe distance of eventual obstacles. Our experiments have shown that MPMP is able to compute trajectories that precisely follow a path produced by a Human driver (distance of 0.15 m in average) while smoothly driving IARA at speeds of up to 32.4 km/h (9 m/s).

## I. INTRODUCTION

An autonomous car is a system composed of a hardware platform and a collection of software modules responsible for tasks that include: localization, mapping, simultaneous localization and mapping (SLAM), path planning, motion planning, traffic-light state detection, high level decision making, etc. We have developed a fully autonomous car named *Intelligent Autonomous Robotic Automobile* (IARA, Figure 1). We have used a Ford Escape Hybrid as hardware platform and designed and built a full set of software modules that allow its autonomous operation in urban traffic.

In a typical mission, IARA is at an initial pose (*origin*), $\mathbf{p}_o = (x_o, y_o, \theta_o)$, and the user defines an end-pose, $\mathbf{p}_e = (x_e, y_e, \theta_e)$, and asks it to go from $\mathbf{p}_o$ to $\mathbf{p}_e$. Once asked to perform such a mission, IARA's path planner builds a path, $P = (\mathbf{pv}_o, \dots, \mathbf{pv}_i, \dots, \mathbf{pv}_e)$, from $\mathbf{p}_o$ to $\mathbf{p}_e$, composed of a vector of poses, about 0.5 m apart, and associated velocities, $\mathbf{pv}_i = (x_i, y_i, \theta_i, v_i)$, that goes from $\mathbf{pv}_o = (x_o, y_o, \theta_o, v_o)$ to $\mathbf{pv}_e = (x_e, y_e, \theta_e, v_e)$, while obeying restrictions imposed by the road limits and the platform's previously defined maximum operating parameters, such as maximum speed, acceleration, rate of driving wheel turn ($d\varphi/dt$), etc. IARA's high level decision making module periodically (20 Hz) slices this path and takes a small portion of it of about 100 m that we call *lane*, or $L = (\mathbf{pv}_1, \dots, \mathbf{pv}_g, \dots, \mathbf{pv}_{|L|})$, and stablish a *goal* in it, $\mathbf{pv}_g = (x_g, y_g, \theta_g, v_g)$, about 5 s in front of IARA's current position.

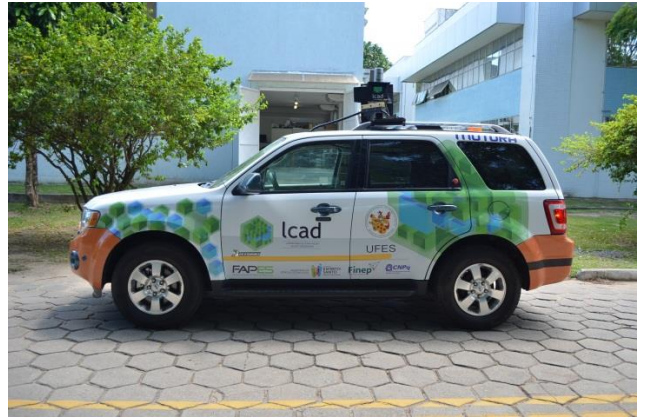

Figure 1: Intelligent Autonomous Robotic Automobile (IARA). A video of IARA's autonomous operation is available at https://goo.gl/RT1EBt.

In this paper, we present IARA's model-predictive motion planning module, which receives a map, a lane and a goal at a rate of 20 Hz, and computes a trajectory, $T = (\mathbf{x}_o, \dots, \mathbf{x}_i, \dots, \mathbf{x}_f)$, from the current car state, $\mathbf{x}_o = (x_o, y_o, \theta_o, v_o, \varphi_o, t_o)$, to a pose and velocity as close as possible to $\mathbf{pv}_g$, while following the lane and avoiding eventual obstacles (Figure 2). The value $\varphi_o$ in $\mathbf{x}_o$ is the front-wheels' steering angle of the current car state (Figure 4).

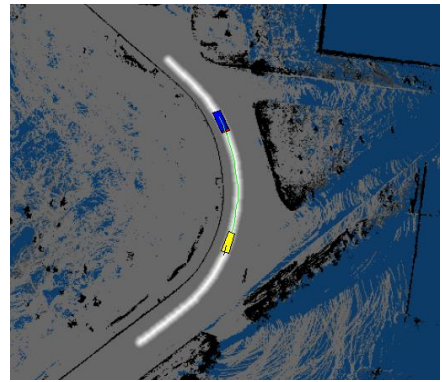

Figure 2: IARA's map, lane and motion plan. IARA is the blue rectangle, while the goal is the yellow rectangle. The lane is show in white and the trajectory $T$ in green.

IARA's model predictive motion planner uses a parameterized model of the behavior of the hardware platform, $M(.)$, that, given an initial state, $\mathbf{x}_o$, the goal, $\mathbf{pv}_g$,

*Research supported by Conselho Nacional de Desenvolvimento Científico e Tecnológico – CNPq, Brazil (grants 552630/2011-0 and 312786/2014-1) and Fundação de Amparo à Pesquisa do Espírito Santo – FAPES, Brazil (grant 48511579/2009).

Vinicius Cardoso, Josias Oliveira, Thomas Teixeira, Claudine Badue, Thiago Oliveira-Santos, Lucas Veronese and Alberto F. De Souza are with Departamento de Informática, Universidade Federal do Espírito Santo, Vitoria, ES, 29075-910, Brazil (phone: +55-27-4009-2138; fax: +55-27-4009-5848; e-mail: {vinicius, josias, thomas, claudine, todsantos, lucas.veronese, alberto}@lcad.inf.ufes.br).

Filipe Mutz is with the Coordenação de Informática, Instituto Federal do Espírito Santo, Vitória, ES, 29040-860, Brazil (email: filipe.mutz@ifes.edu.br).

and a set of *trajectory control parameters*, **tcp**, computes the trajectory $T = M(\mathbf{x}_o, \mathbf{pv}_g, \mathbf{tcp})$. In order to obtain $T$, we have to find the right **tcp**. For that, using the conjugate gradient optimization algorithm, we gradually change an initial **tcp** seed taken from a pre-computed table indexed by indexes derived from $\mathbf{x}_o$ and $\mathbf{pv}_g$, so that: (i) $\mathbf{pv}_f = (x_f, y_f, \theta_f, v_f)$, composed of the final pose and velocity at state $\mathbf{x}_f$, is close enough to $\mathbf{pv}_g$ according to some optimization criteria; (ii) the poses in $T$ are as close as possible to the poses in $L$; and (iii) the poses in $T$ are as far as possible of eventual obstacles in the given map, $map$.

We have evaluated IARA's model-predictive motion planner (MPMP) experimentally by comparing it with a Human driver. Our results show that MPMP performance compares well with Human performance – its path is smooth and very close to the Human path (average distance of 0.15 m, $\sigma = 0.14$) and its speeds are more stable than that of the Human driver. Currently, MPMP can safely navigate IARA in urban environments with speeds of up to 32.4 km/h (9 m/s).

## II. RELATED WORK

There are various methods in the literature to address the problem of *on-road motion planning*[1] for autonomous cars. Readers are referred to Bautista et al. [1] for a review on these methods. Among those that were evaluated experimentally using *real-world autonomous cars*, on-road motion planning methods employ mainly state lattice, rapidly-exploring random tree (RRT), interpolation, optimization, and model predictive techniques.

In methods based on state lattice [2] [3], trajectories between initial and desired goal states are searched for in a state lattice that is adapted for on road motion planning, such that only states that are a priori likely to be in the solution are represented. Possible trajectories are evaluated by a cost function that considers the car's dynamic, environmental and behavioral constraints, among others. The major disadvantage of these methods is that they are computationally costly due to the evaluation of every possible solution in the graph.

In methods based on RRT [4] [5], a search tree is built incrementally from the car's initial state using random states. For each random state, a control command is applied to the nearest state of the tree for creating a new state as close as possible to the random state. A trajectory is found when a state reaches the goal state. Candidate trajectories are evaluated according to various criteria. The main weakness of these methods is that solutions are not continuous and, therefore, jerky.

In methods based on optimization [6] [7], trajectories are computed using an optimization process, which is run over trajectory parameters, and aims at minimizing an objective function that considers trajectory constraints, such as position, velocity, acceleration and jerk. The shortcoming of these methods is that they are time consuming, since the optimization process takes place at each motion state.

---

[1] We refer to the problem of *motion planning* that aims at planning a trajectory – list of commands of velocity and steering angle along with respective execution times – considering kinematic and dynamic constraints of the autonomous car, which contrast to the problem of *path planning*, that aims at planning a path – list of waypoints – considering only kinematic constraints. We also refer to the problem of *on-road motion planning* that aims at planning a trajectory that follows a desired lane, which differs from the problem of *unstructured motion planning* in which there are no lanes and, thus, the trajectory is far less constrained.

In model predictive methods [8], a model predictive trajectory generation algorithm is used to compute trajectories that satisfy desired goal state constraints. Trajectories are computed via an optimization procedure that is run over control parameters. The constraint equation is defined as the difference between the goal state constraints and the integral of the car's model dynamics. The constrained trajectory generation algorithm determines the control parameters that minimize the constraint equation. An approximate mapping from the state space to the parameter space is precomputed offline, in order to seed the constrained optimization process. Resulting trajectories are evaluated using several criteria.

Amidst previous works cited above, the most similar to ours is the motion planner proposed by Ferguson et al. [8]. However, our work differs from that of Ferguson et al. in three main aspects. First, our planner is able to compute more complex trajectories, since it uses a steering angle spline with one extra knot point (four in total) to parameterize the car control, while that of Ferguson et al. uses a three knot curvature spline. Second, our planner can generate trajectories that are optimized for curvy roads with obstacles, since its cost function considers the desired road as well as obstacles, while that of Ferguson et al. generates several alternative trajectories and selects a collision-free one. Third, we present the algorithm that we use for computing the table of **tcp** seeds.

## III. IARA'S MODEL-PREDICTIVE MOTION PLANNER

### A. Hardware Platform Model

IARA's model-predictive motion planner uses a parameterized model of the behavior of the hardware platform, $M(.)$, that, given an initial state, $\mathbf{x}_o = (x_o, y_o, \theta_o, v_o, \varphi_o, t_o)$, the goal, $\mathbf{pv}_g = (x_g, y_g, \theta_g, v_g)$, and a set of trajectory control parameters, **tcp**, computes a trajectory $T = M(\mathbf{x}_o, \mathbf{pv}_g, \mathbf{tcp})$, which is a vector of states, $T = (\mathbf{x}_o, \ldots, \mathbf{x}_i, \ldots, \mathbf{x}_f)$. The elements of **tcp** are the total time of the trajectory, $tt$, and three knot points, $k_1$, $k_2$ and $k_3$, $\mathbf{tcp} = (tt, k_1, k_2, k_3)$. The three knot points, together with $\varphi_o$, define a cubic spline that specifies the evolution of the steering angle during $tt$ (Figure 3).

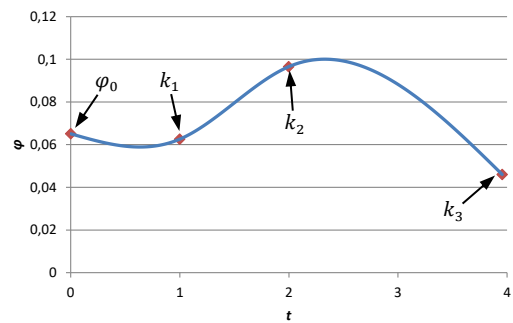

Figure 3: Cubic spline, defined by $\varphi_o$, and $k_1$, $k_2$ and $k_3$, that specifies the evolution of the steering angle during $tt$, for $tt = 3.96$ s. The knot point $k_1$ is defined at $t = tt/4$, $k_2$ at $t = tt/2$ and $k_3$ at $t = tt$.

Our car model, $M(.)$, is currently a bicycle kinematic model (Figure 4) modified to consider an understeer coefficient, $u$, and is defined by the equations:

$$x_{t+1} = x_t + \Delta t\, v_t \cos \theta_t, \quad (1)$$
$$y_{t+1} = y_t + \Delta t\, v_t \sin \theta_t, \quad (2)$$
$$\theta_{t+1} = \theta_t + \Delta t\, v_t c_t, \quad (3)$$
$$v_{t+1} = v_t + \Delta t\, a, \quad (4)$$
$$\varphi_{t+1} = spline(\Delta t(t+1)), \text{ and} \quad (5)$$
$$c_t = \frac{\tan \frac{\varphi_t}{1+uv_t^2}}{l} \quad (6)$$

where $c_t$ is the car curvature that is directly used to control the car [9]. The relationship between $\varphi_t$ and $c_t$, given by Equation (6), together with equations (1) to (5), constitute a simplification of the full Ackerman car model and it was obtained experimentally [10]. For our hardware platform, the understeer coefficient, $u$, is equal to 0.0015 [9]. This car model can be improved in the future without a significant impact on the remainder of the planner.

To obtain $T$, $M(.)$ is interactively used for computing each state, $\mathbf{x}_{t+1}$. During each planning cycle, the model starts with $t = 0$ and $a = (v_g - v_0)/tt$, and it is run from $t = 0$ to $t = tt/\Delta t$.

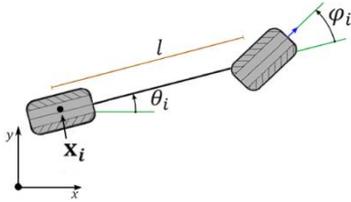

Figure 4: Bicycle kinematic model (figure adapted from [11])

*B. Minimization Problem*

At a rate of 20 Hz, our model-predictive planner computes a trajectory, $T$, that starts at the current car state, $\mathbf{x}_o = (x_o, y_o, \theta_o, v_o, \varphi_o, t_o)$, and finishes at a state, $\mathbf{x}_f = (x_f, y_f, \theta_f, v_f, \varphi_f, t_f)$, whose pose and velocity, $\mathbf{pv}_f = (x_f, y_f, \theta_f, v_f)$, is as close as possible to a given goal pose and velocity, $\mathbf{pv}_g = (x_g, y_g, \theta_g, v_g)$. However, it has to do that while keeping the car at a safe distance from obstacles and as close as possible to the poses of the given lane $L = (\mathbf{pv}_1, \cdots, \mathbf{pv}_g, \cdots, \mathbf{pv}_{|L|})$, part of the path computed by the IARA's path planner. To compute $T$, we solve the following minimization problem:

$$\arg\min_{\mathbf{tcp}} f(M(\mathbf{x}_o, \mathbf{pv}_g, \mathbf{tcp}), L, map) \quad (7)$$

where

$$f(.) = \sqrt[2]{w_1 \Delta\lambda^2 + w_2 \Delta\theta^2 + w_3 \Delta\phi^2 + w_4 D^{O^2} + w_5 D^{L^2}}, \quad (8)$$

and $w_1$ to $w_5$ are weights.

To minimize $f(.)$, we have to find the **tcp** that minimizes the square root of the weighted sum of the squares of the values: (i) $\Delta\lambda$, (ii) $\Delta\theta$ and (iii) $\Delta\phi$, which, together, measure the distance from the end of $T$, $\mathbf{pv}_f$, to the desired goal, $\mathbf{pv}_g$; (iv) $D^O$, which summarizes the distance of each point in $T$ to its nearest obstacle; and (v) $D^L$, which summarizes the distances between each point in $L$ and its nearest point in $T$.

The value $\Delta\lambda$ is the difference between the magnitude of the two vectors, $\mathbf{s}_g = (x_g, y_g)$ and $\mathbf{s}_f = (x_f, y_f)$, that connect the poses associated with $\mathbf{x}_o$ and $\mathbf{pv}_g$ ($\mathbf{s}_g$), and the poses associated with $\mathbf{x}_o$ and $\mathbf{pv}_f$ ($\mathbf{s}_f$), and it is computed by the equation (Figure 5):

$$\Delta\lambda = \|(x_g - x_0, y_g - y_0)\| \\ - \|(x_f - x_0, y_f - y_0)\| \quad (9)$$

The value $\Delta\theta$ is the difference between the car orientation at the goal and the car orientation at the end of the trajectory, and is computed by the equation (Figure 5):

$$\Delta\theta = \theta_g - \theta_f \quad (10)$$

The value $\Delta\phi$ is the angle between $\mathbf{s}_g$ and $\mathbf{s}_f$, and is given by the equation (Figure 5):

$$\Delta\phi = \text{atan2}((y_g - y_o)/(x_g - x_o)) \\ - \text{atan2}((y_f - y_o)/(x_f - x_o)) \quad (11)$$

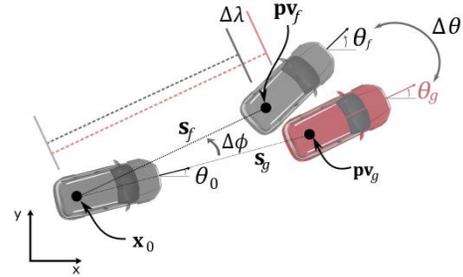

Figure 5: Variables associated with $\Delta\lambda$, $\Delta\theta$ and $\Delta\phi$, which, together, measure the distance from the end of the $T$, $\mathbf{pv}_f$, to the desired goal, $\mathbf{pv}_g$.

The value $D^O$ is the summation of the differences between $d^{min}$ and $\|\mathbf{x}_i - \mathbf{o}_i\|$, where $d^{min}$ is the smallest allowed distance between the car and an obstacle (about half the car width), and $\|\mathbf{x}_i - \mathbf{o}_i\|$ is the distance between the pose of $\mathbf{x}_i \in T$ and its nearest obstacle, $\mathbf{o}_i$, in the occupancy grid map, $map$, and is computed by the equation:

$$D^O = \sum_{i=0}^{f} \max\left(0, d^{min} - \|\mathbf{x}_i - \mathbf{o}_i\|\right) \quad (12)$$

Finally, the value $D^L$ is the summation of $\|\mathbf{pv}_k - \mathbf{x}_i\|$ from $\mathbf{pv}_j$ (the pose in $L$ nearest to $\mathbf{x}_o$) to $\mathbf{pv}_g$ (the goal pose in $L$), where $\|\mathbf{pv}_k - \mathbf{x}_i\|$ is distance between the pose of each element $\mathbf{pv}_k$ of $L$ to the pose of $\mathbf{x}_i \in T$ nearest to $\mathbf{pv}_k$, and is calculated by the equation (Figure 6):

$$D^L = \sum_{k=j}^{g} \|\mathbf{pv}_k - \mathbf{x}_i\| \quad (13)$$

To solve the minimization problem, we start with an initial guess for **tcp**, $\mathbf{tcp}^{seed}$, taken from a 5-dimension pre-computed table, named *trajectory look-up table*, $TLT$, which is indexed by indexes computed from $\mathbf{x}_o$ and $\mathbf{pv}_g$, named *discrete trajectory descriptors*, $\mathbf{dtd} = (\lambda_g^d, \phi_g^d, \theta_g^d, v_o^d, \varphi_o^d)$, as described in Section III.D below. The indexes $\lambda_g^d$, $\phi_g^d$, $\theta_g^d$, $v_o^d$ and $\varphi_o^d$ are computed from $\lambda_g$, $\phi_g$, $\theta_g$, $v_o$ and $\varphi_o$, respectively, which, as a tuple, are called *trajectory descriptors*, or $\mathbf{td} = (\lambda_g, \phi_g, \theta_g, v_o, \varphi_o)$ (Section III.D). We then use the conjugate gradient optimization algorithm to minimize $f(.)$ by manipulating the $\mathbf{tcp}^{seed}$ elements (the trajectory total time, $tt$, and three knot points, $k_1$, $k_2$ and $k_3$, of the cubic spline that specifies the evolution of the steering angle during $tt$).

The conjugate gradient algorithm requires derivatives of the cost function, $f(.)$, with respect to the optimizing variables, $tt$, $k_1$, $k_2$ and $k_3$. In our model-predictive planner, we compute these derivatives numerically using finite differences.

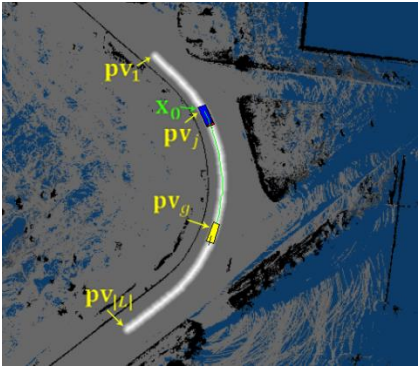

Figure 6: Variables associated with $D^L$, which summarizes the distances between each point in $L$ and its nearest point in $T$.

## C. Model-Predictive Planner Algorithm

The listing of Algorithm 1 presents our model-predictive motion planning algorithm. At a rate of 20 Hz, compute_motion_plan() receives $\mathbf{x}_o$, $\mathbf{pv}_g$, $L$ and $map$ from IARA's path planner as input, and returns $T$ as output.

**Algorithm 1:** compute_motion_plan($\mathbf{x}_o, \mathbf{pv}_g, L, map$)
1: $\mathbf{td} \leftarrow$ get_td($\mathbf{x}_o, \mathbf{pv}_g$)
2: $\mathbf{dtd} \leftarrow$ get_dtd($\mathbf{td}$)
3: $\mathbf{tcp}^{seed} \leftarrow TLT\left[\mathbf{dtd}^{(\lambda_g^d)}\right]\left[\mathbf{dtd}^{(\phi_g^d)}\right]\left[\mathbf{dtd}^{(\theta_g^d)}\right]\left[\mathbf{dtd}^{(\varphi_o^d)}\right]\left[\mathbf{dtd}^{(v_o^d)}\right]$
4: $\mathbf{tcp}^{optimized} \leftarrow \arg\min_{\mathbf{tcp}} f(M(\mathbf{x}_o, \mathbf{pv}_g, \mathbf{tcp}^{seed}), L, map)$
5: **if** valid($\mathbf{tcp}^{optimized}$) **then**
6: $\quad a \leftarrow (\mathbf{pv}_g^{(v_g)} - \mathbf{x}_o^{(v_o)})/\mathbf{tcp}^{optimized\,(tt)}$
7: $\quad T \leftarrow$ get_T_by_simulation($\mathbf{x}_o, a, \mathbf{tcp}^{optimized}$)
8: **else**
9: $\quad T \leftarrow \emptyset$
10: **return** ($T$)

In line 1, the Algorithm 1 uses the function get_td() to compute $\mathbf{td}$ as a function of $\mathbf{x}_o$ and $\mathbf{pv}_g$ (Section III.D). In line 2, it uses the function get_dtd() to compute $\mathbf{dtd}$ from $\mathbf{td}$ (Section III.D). In line 3, it gets $\mathbf{tcp}^{seed}$ from $TLT$ and, in line 4, it runs the conjugate gradient algorithm to find a $\mathbf{tcp}^{optimized}$ that takes the car from $\mathbf{x}_o$ to $\mathbf{pv}_g$. In line 6, the acceleration $a$ is computed. If the optimization succeeds, in line 7, the function get_T_by_simulation() (trivial implementation; listing not shown) simulates the car, according to equations (1) to (6), starting from $\mathbf{x}_o$ (time $t = 0$) and until time $tt$ of $\mathbf{tcp}^{optimized}$ using $a$. At the end of this simulation, the function get_T_by_simulation() returns a trajectory $T$ that ends as close as possible to $\mathbf{pv}_g$ and with poses: (i) as close as possible to $L$ and (ii) as far as possible from the obstacles in $map$. If the optimization fails, in line 9 an empty trajectory is returned in the cycle. If compute_motion_plan() fails to find trajectories for many consecutive cycles, IARA is stopped. Note that IARA does not move much in a single cycle and a previous trajectory might still be suitable after several cycles. IARA's obstacle avoider module takes care of it in such cycles (or in any situation that may lead to a collision) [12].

## D. Trajectory Look-up Table – TLT

Initial guesses for $\mathbf{tcp}$, $\mathbf{tcp}^{seed}$, are pre-computed with varying $\mathbf{td} = (\lambda_g, \phi_g, \theta_g, v_o, \varphi_o)$, where $\lambda_g$ and $\phi_g$ are the relative differences between $\mathbf{x}_o$ and $\mathbf{pv}_g$ in polar coordinates; $\theta_g$ is the goal orientation; $v_o$ is the initial velocity; and $\varphi_o$ is the initial steering angle (Figure 7). A $\mathbf{tcp}^{seed}$ computed from $\mathbf{td} = (\lambda_g, \phi_g, \theta_g, v_o, \varphi_o)$ is stored in the $TLT$ cell indexed by $\mathbf{dtd} = (\lambda_g^d, \phi_g^d, \theta_g^d, v_o^d, \varphi_o^d)$, i.e., $TLT[\lambda_g^d][\phi_g^d][\theta_g^d][v_o^d][\varphi_o^d] = \mathbf{tcp}^{seed}$. The elements of $\mathbf{td}$ are derived from $\mathbf{x}_o$ and $\mathbf{pv}_g$ according to the following equations:

$$\lambda_g = \|(x_g - x_o, y_g - y_o)\|, \quad (14)$$
$$\phi_g = \mathrm{atan2}((y_g - y_o)/(x_g - x_o)), \quad (15)$$
$$\theta_g = \mathbf{pv}_g^{(\theta)}, \quad (16)$$
$$v_o = \mathbf{x}_o^{(v)} \text{ and} \quad (17)$$
$$\varphi_o = \mathbf{x}_o^{(\varphi)}, \quad (18)$$

where, in the equations (16), (17) and (18) above, $\mathbf{tuple}^{(a)}$ is the element $a$ of a $\mathbf{tuple} = (a, b, c, \dots)$.

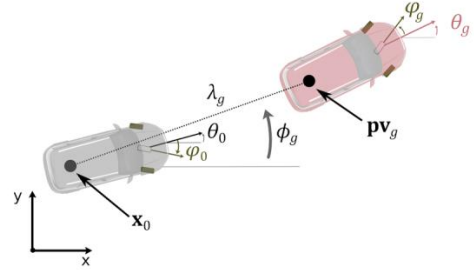

Figure 7: Elements of a $\mathbf{td}$ ($\lambda_g$, $\phi_g$, $\theta_g$, $v_o$ and $\varphi_o$), which are derived from $\mathbf{x}_o$ and $\mathbf{pv}_g$

The discrete indexes, $\lambda_g^d$, $\phi_g^d$, $\theta_g^d$, $v_o^d$ and $\varphi_o^d$, are defined as:

$$\lambda_g^d = g(\lambda_g, cr^{\lambda_g}, sf^{\lambda_g}, zi^{\lambda_g}), \quad (19)$$
$$\phi_g^d = h(\phi_g, cd^{\phi_g}, zi^{\phi_g}), \quad (20)$$
$$\theta_g^d = g(\theta_g, cr^{\theta_g}, sf^{\theta_g}, zi^{\theta_g}), \quad (21)$$
$$v_o^d = g(v_o, cr^{v_o}, sf^{v_o}, zi^{v_o}) \text{ and} \quad (22)$$
$$\varphi_o^d = g(\varphi_o, cr^{\varphi_o}, sf^{\varphi_o}, zi^{\varphi_o}), \quad (23)$$

where

$$g(\iota, cr, sf, zi) = \left[\log_{cr} \frac{\iota + sf}{sf}\right] + zi \quad (24)$$

for $\iota = \lambda_g$ (or $\theta_g$, $v_o$ and $\varphi_o$), $cr = cr^{\lambda_g}$ (or $cr^{\theta_g}$, $cr^{v_o}$ and $cr^{\varphi_o}$), $sf = sf^{\lambda_g}$ (or $sf^{\theta_g}$, $sf^{v_o}$ and $sf^{\varphi_o}$) and $zi = zi^{\lambda_g}$ (or $zi^{\theta_g}$, $zi^{v_o}$ and $zi^{\varphi_o}$);

$$h(\iota, cd, zi) = \left[\frac{\iota}{cd}\right] + zi \quad (25)$$

for $\iota = \phi_g$, $cd = cd^{\phi_g}$ and $zi = zi^{\phi_g}$; and, in the equations (24) and (25), $[e]$ is equal to the nearest integer to $e$.

The functions $g(.)$ and $h(.)$, and their parameters for each element of **dtd**, were chosen in order to compute a **dtd** suitable for accessing a $TLT$ that would contain the appropriate **tcp**$^{seed}$ for building trajectories with properties in the range of operation of IARA. More specifically, the functions $g(.)$ and $h(.)$ were chosen in order to apply finer-grained discretizations of the elements of **dtd** for smaller relative differences between $\mathbf{x}_o$ and $\mathbf{pv}_g$ and coarser-grained discretizations for larger ones. Table I presents the values we have chosen for the parameters of $g(.)$ and $h(.)$ of each element of **dtd**, and the bounds, $n$, of $g(.)$ and $h(.)$ for each element. We also have functions $g^{-1}(.)$ and $h^{-1}(.)$ that compute the inverse of $g(.)$ and $h(.)$, i.e., map each member of **dtd** to an element of **td** (trivial deduction; not shown).

TABLE I: $g(.)$ AND $h(.)$ PARAMETERS AND BOUNDS

| dtd | $g(.)$ and $h(.)$ parameters | | | | bounds ($n$) | |
| --- | --- | --- | --- | --- | --- | --- |
| | cr | sf | zi | cd | min | Max |
| $\lambda_g^d$ | 1.8 | 2.3 | -1 | - | 0 | 15 |
| $\phi_g^d$ | - | - | 7 | 0.139 | 0 | 15 |
| $\theta_g^d$ | 1.3 | 0.174 | 7 | - | 0 | 15 |
| $\varphi_o^d$ | 1.394 | 0.052 | 7 | - | 0 | 15 |
| $v_o^d$ | 1.381 | 1.3 | 0 | - | 0 | 8 |

The listing of Algorithm 2 presents the algorithm designed to generate the $TLT$. In line 1, Algorithm 2 initializes $TLT$, $L$ and $map$ as empty (when building a **tcp**$^{seed}$ one does not need to consider a lane or $map$). In lines 2, 3 and 4, Algorithm 2 cycles through all possible values of $\lambda_g^d$, $v_o^d$ and $\varphi_o^d$ (i.e. within the bounds of Table I), while, in lines 5 and 6, it samples $k_2$ and $k_3$ throughout the range of possible IARA's steering angles values (at small intervals). The value $min^\varphi$ is the minimum steering angle and $max^\varphi$ is the maximum steering angle. Throughout lines 7 to 14, Algorithm 2 use all these values to compute a **tcp**$^{seed}$, which is saved into $TLT$ in line 16.

**Algorithm 2:** fill_trajectory_lookup_table()
1: $TLT \leftarrow \emptyset, L \leftarrow \emptyset, map \leftarrow \emptyset$
2: **for** $\lambda_g^d = 0$ **to** $n^{\lambda_g} - 1$
3:  **for** $v_o^d = 0$ **to** $n^{v_o} - 1$
4:   **for** $\varphi_o^d = 0$ **to** $n^{\varphi_o} - 1$
5:    **for** $k_2 = min^\varphi$ **to** $max^\varphi$
6:     **for** $k_3 = min^\varphi$ **to** $max^\varphi$
7:      $\mathbf{tcp}^{sample}, a \leftarrow$ get_tcp_sample_and_acc($\lambda_g^d, v_o^d, k_2, k_3$)
8:      $\mathbf{x}_o = (0,0,0, g^{-1}(v_o^d), g^{-1}(\varphi_o^d), 0)$
9:      $\mathbf{td}, v_g \leftarrow$ get_td_and_vg_by_simulation($\mathbf{x}_o, a, \mathbf{tcp}^{sample}$)
10:     $\mathbf{dtd} \leftarrow$ get_dtd($\mathbf{td}$)
11:     **if** valid(**dtd**) **then**
12:      $\lambda = g^{-1}(\mathbf{dtd}^{(\lambda_g^d)}), \phi = h^{-1}(\mathbf{dtd}^{(\phi_g^d)}), \theta = g^{-1}(\mathbf{dtd}^{(\theta_g^d)})$
13:      $\mathbf{pv}_g = (\lambda \cos\phi, \lambda \sin\phi, \theta, v_g)$
14:      $\mathbf{tcp}^{seed} \leftarrow \arg\min_{\mathbf{tcp}} f(M(\mathbf{x}_o, \mathbf{pv}_g, \mathbf{tcp}^{sample}), L, map)$
15:      **if** valid($\mathbf{tcp}^{seed}$) **then**
16:       $TLT\left[\mathbf{dtd}^{(\lambda_g^d)}\right]\left[\mathbf{dtd}^{(\phi_g^d)}\right]\left[\mathbf{dtd}^{(\theta_g^d)}\right]\left[\mathbf{dtd}^{(\varphi_o^d)}\right]\left[\mathbf{dtd}^{(v_o^d)}\right] \leftarrow \mathbf{tcp}^{seed}$
17: **return** ($TLT$)

The function get_tcp_sample_and_acc(), called in line 7 of Algorithm 2, is presented in Algorithm 3. This function computes a **tcp**$^{sample}$ and an acceleration, $a$. The elements $k_2$ and $k_3$ of **tcp**$^{sample}$ come directly from the inputs of get_tcp_sample_and_acc(), while $tt$ of **tcp**$^{sample}$ is computed in lines 3 to 9 of Algorithm 3. Note that $k_1$ is not filled in **tcp**$^{seed}$, since this part of a **tcp** is only computed during run time – $k_1$ is used for adding maneuverability to avoid obstacles and to keep IARA precisely in the road during the execution of compute_motion_plan(), Algorithm 1. Finally, the acceleration $a$ is computed in line 10 of Algorithm 3 using the equation of motion, $\lambda = v_o \times t + a \times t^2/2$, solved for the acceleration ($a = (\lambda - v_o \times t)/(t^2/2)$).

**Algorithm 3:** get_tcp_sample_and_acc($\lambda_g^d, v_o^d, k_2, k_3$)
1: $\mathbf{tcp}^{sample(k_2)} = k_2$
2: $\mathbf{tcp}^{sample(k_3)} = k_3$
3: $\lambda = g^{-1}(\lambda_g^d)$
4: **if** $\lambda > 7$ **then**
5:  $\mathbf{tcp}^{sample(tt)} = 5s$
6: **else if** $\lambda > 3.5$ **then**
7:  $\mathbf{tcp}^{sample(tt)} = 2.5s$
8: **else**
9:  $\mathbf{tcp}^{sample(tt)} = 2s$
10: $a = (\lambda - g^{-1}(v_o^d) \times tt)/(tt^2/2)$
11: **return** ($\mathbf{tcp}^{sample}, a$)

Algorithm 2 initializes the car at the origin in line 8, but with velocity $g^{-1}(v_o^d)$ and steering angle $g^{-1}(\varphi_o^d)$. In line 9, using the function get_td_and_vg_by_simulation() (trivial implementation; listing not shown), it simulates the car, according to equations (1) to (6), from $\mathbf{x}_o$ (time $t = 0$) until time $tt$ of **tcp**$^{sample}$ using acceleration $a$. At the end of this simulation, get_td_and_vg_by_simulation() computes a **td** using the last state of the simulation as goal. The function get_td_and_vg_by_simulation() then returns this **td** as well as the velocity of the last state of the simulation as $v_g$.

In line 10, Algorithm 2 uses the function get_dtd() (trivial implementation; listing not shown) to compute a **dtd** from a **td**, according to equations (19) to (23). This **dtd** might be invalid (one of its elements might be out of the bounds shown in Table I) and, in such cases, it is discarded. If this **dtd** is valid, in lines 12 and 13, Algorithm 2 builds a $\mathbf{pv}_g$ and, in line 14, runs the conjugate gradient algorithm to find a **tcp** that takes the car from $\mathbf{x}_o$ (time $t = 0$) to $\mathbf{pv}_g$ (time $t = tt$). If the optimization succeeds, in line 16 the computed **tcp**$^{seed}$ is saved in $TLT$.

Many combinations of $\lambda_g^d, \phi_g^d, \theta_g^d, \varphi_o^d, v_o^d$ can never occur in real life due to the physical restrictions of IARA. For example, it cannot change from an angle $\theta_o$ to a largely different angle $\theta_g$ in a short $\lambda_g$. Therefore, many $TLT$ cells are empty after Algorithm 2. To fill as much cells as possible, after Algorithm 2, we run a function that goes back to each empty cell of $TLT$ and use nearby occupied cells as seeds for computing the **tcp** of these empty cells (listing not shown). This process is repeated until no extra cell is filled with a valid **tcp**. Using this process, we were able to fill 57.63% of $TLT$. Our experimental evaluation has shown that this is enough for proper run time operation.

## IV. EXPERIMENTAL SETUP

### A. IARA's Hardware and Software

We developed the hardware and software of IARA (Figure 1). IARA's hardware is based on a Ford Escape Hybrid, which was adapted by Torc Robotics (http://www.torcrobotics.com) to enable electronic actuation of the steering wheel, throttle and brake; reading the car odometry; and powering several high-performance sensors and computers. The IARA's hardware includes one Velodyne HDL 32-E Light Detection and Ranging (LiDAR); one Trimble Dual Antenna RTK GPS; one Xsens Mti IMU; four Point Grey Bumblebee stereo cameras; and two Dell Precision R5500 computers.

The IARA's software is composed of six main modules: *behavior selector*, *localizer* [13] [14], *mapper* [15], *path planner*, *motion planner* (the focus of this paper) and *obstacle avoider* [12]. Together they allow IARA's autonomous operation in urban roads. Additional modules include: *health monitor*, *logger*, *traffic lights state detector*, *simulator* [16], among others.

### B. Test Environment

IARA's test environment is the *Universidade Federal do Espírito Santo* (www.ufes.br) main campus beltway (UFES beltway). Figure 8 shows the 3.7 km of the UFES beltway. It is a very challenging course since it is frequently busy, with many cars, motorcycles and busses travelling around, parked cars on both sides of the road, as well as pedestrians crossing (Figure 8(2)). The road pavement includes segments with cobbles (Figure 8(4)) and asphalt (Figure 8(3)). Furthermore, it has six speed bumps, sharp and wide curves, varying track widths and two gate barriers. Therefore, the motion planner must be able to deal with all these hazards in order to drive the car smoothly and safely.

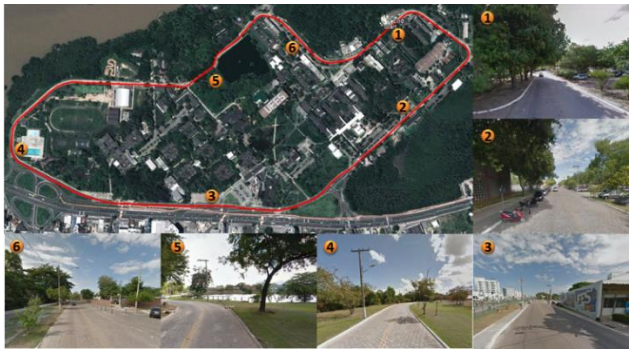

Figure 8: UFES's beltway [15]. Google-Earth and Google-StreetView pictures of the university testing enviroment. The top left image shows the 3.7 km course in red. The images 1 to 6 show some of the relevant parts of the course.

### C. Test Methodology

We asked an experienced Human driver to drive IARA along the UFES beltway and logged the car poses (given by our localizer module). We transformed the sequence of poses and velocities observed during Human driving into a standard IARA path (as would be produced by our path planner). Later, we asked IARA to perform the same path in autonomous mode using the MPMP. We evaluated the differences between the Human path and the MPMP path.

Furthermore, we show the behavior of MPMP with obstacles. The results of these evaluations are reported below.

## V. EXPERIMENTAL EVALUATION

Figure 9 compares the performance of MPMP driving with that of Human driving. Figure 9(a) shows the distance between each point of the path followed by MPMP to the path followed by the Human driver. In the graph of this figure, in the $x$ axis we show the time of each pose of the Human path, while in the $y$ axis we show the distance between paths. Note that we have synchronized the poses of the MPMP path with the poses of the Human path so as to measure the distance between the correct poses, since the velocities were not the same in both paths. As Figure 9(a) shows, the absolute distance between paths is small throughout the whole paths, never exceeding 0.8 m. The average distance was 0.15 m ($\sigma = 0.14$).

Figure 9(b) compares the velocities of MPMP and Human along the UFES beltway. In the graph of this figure, in the $x$ axis we show the time of each velocity sample, while in the $y$ axis we show the velocities (the velocity samples were synchronized with the time of the poses of Figure 9(a)). The two curves in the graph show the MPMP velocity in red and the Human velocity in blue.

The maximum allowed speed in the UFES beltway varies, but, in most of the beltway, it is 30 km/h (8.33 m/s). We programmed MPMP to maintain 30 km/h (8.33 m/s) and asked the Human driver to follow the beltway speed limits. As the graph in Figure 9(b) shows, MPMP managed to maintain IARA's speed close to the programmed velocity. Its speed was reduced to less than 7.2 km/h (2 m/s) in some points due to the bumps and gate barriers present in the beltway. As can be seen in Figure 9(b), MPMP was more cautious while crossing bumps, gates barriers and other places that required slower speeds. Also, its speed was more stable than the Human speed throughout the course.

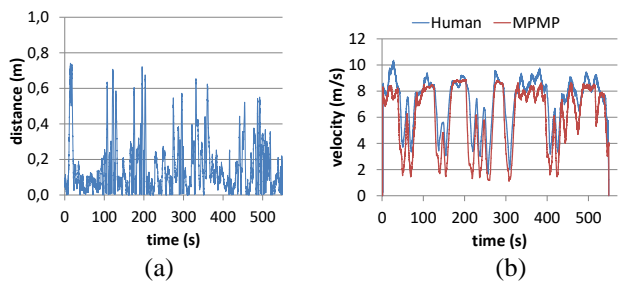

(a)             (b)

Figure 9: The performance of the MPMP driving IARA in a real world environment. (a) Absolute distance between MPMP poses and Human poses. (b) MPMP velocities (red) and Human velocities (blue).

Figure 10 compares the poses of Human (blue) and MPMP (red) throughout the UFES beltway (please compare Figure 8 with Figure 10). As Figure 10 shows, in the scale of the whole beltway, the poses of the two drivers are indistinguishable. The inset in Figure 10 highlights one of the regions of the beltway were the distance between the paths are around the largest. As this inset shows, both paths are smooth and the distance between them changes smoothly as well.

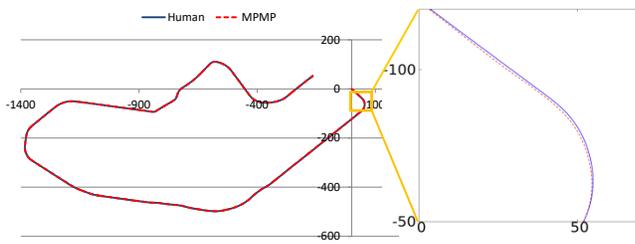

Figure 10: Poses of the two paths throughout the UFES beltway.

In real traffic situations, the planner has also to deal with obstacles in the road. Figure 11 shows one such situation that happened in the UFES beltway. Several busses were parked on the side the road, blocking part of it. The MPMP had to optimize the trajectory to maintain a safe distance from obstacles while maintaining the trajectory as close as possible to the lane (Equation (9)). Please note that the Brazilian traffic regulations allow crossing school busses (or other cars) in the situation shown in Figure 11.

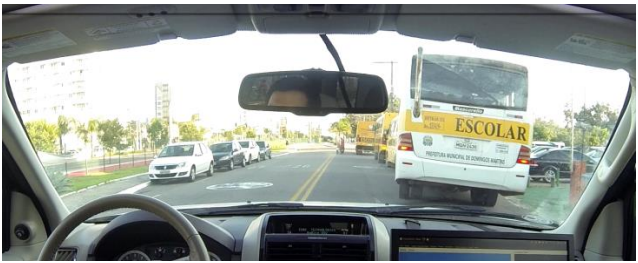

Figure 11: Busses blocking part of the road in UFES beltway.

Figure 12 presents the situation shown in Figure 11 according to the MPMP perspective. The optimized trajectory (green/red) avoids the obstacles between the IARA (rectangle in the beginning of the trajectory, on the right of it) and the goals (yellow rectangle on the left). The MPMP tries to maintain the IARA on the lane, while keeping a safe distance from the obstacles. This experiment can be seen in the video https://youtu.be/o_NU23fpZhw?t=134.

## VI. CONCLUSIONS AND FUTURE WORK

In this work, we have presented the Model-Predictive Motion Planner (MPMP) developed for the Intelligent Autonomous Robotic Automobile (IARA). MPMP is a high frequency motion planner that operates at 20 Hz and is capable of generating smooth trajectories that follow a reference path while avoiding eventual obstacles. We have tested MPMP running in IARA in a challenging course of 3.7 km and compared its performance with that of a Human in the same course (we asked IARA to follow the Human path as close as possible). Our results showed that MPMP compares well with the Human performance – its path is smooth, very close to the Human path (average distance of 0.15 m, $\sigma = 0.14$) and its speeds are more stable than that of the Human driver. Besides, MPMP trajectories obey the restrictions imposed by obstacles in the road and the platform's performance limits (speed, acceleration, rate of driving wheel turn, etc.). Currently, MPMP can safely navigate in urban environments with speeds of up to 32.4 km/h (9 m/s), while performing very close to Human drive behavior.

In future work, we will investigate how to improve MPMP and IARA's low level control to improve its maximum speed. Also, we will examine how to incorporate constraints associated with moving obstacles in our trajectory optimization model and model-predictive motion planning algorithm.

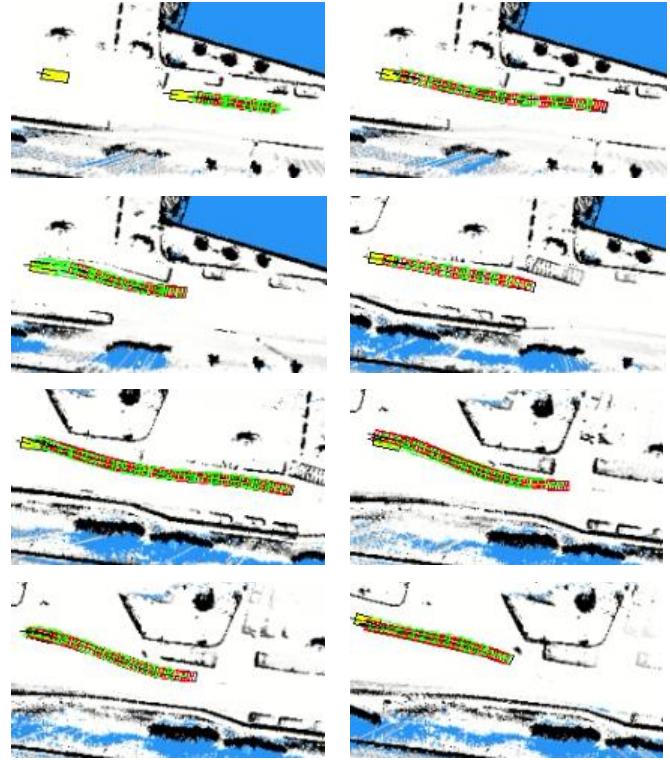

Figure 12: MPMP trajectories (in green/red) avoiding obstacles on road.